\def\year{2020}\relax
\documentclass[letterpaper]{article} % DO NOT CHANGE THIS
\usepackage{aaai20}  % DO NOT CHANGE THIS
\usepackage{times}  % DO NOT CHANGE THIS
\usepackage{helvet} % DO NOT CHANGE THIS
\usepackage{courier}  % DO NOT CHANGE THIS
\usepackage[hyphens]{url}  % DO NOT CHANGE THIS
\usepackage{graphicx} % DO NOT CHANGE THIS
\usepackage{graphicx}  % DO NOT CHANGE THIS
\usepackage{makecell}
\usepackage{comment}
\usepackage[title]{appendix}
\title{Vehicle Emissions Prediction with Physics-Aware AI Models: Preliminary Results}
\author{Harish Panneer Selvam\textsuperscript{\rm 1}, Yan Li\textsuperscript{\rm 2}, Pengyue Wang,\textsuperscript{\rm 1}
\\  \Large  
\textbf{William F. Northrop, \textsuperscript{\rm 1} Shashi Shekhar\textsuperscript{\rm 2}  }\\
\textsuperscript{\rm 1}Dept. of Mechanical Engineering
\textsuperscript{\rm 2}Dept. of Computer Science, University of Minnesota Twin-Cities \\
Minneapolis, Minnesota 55455
\\\{panne027, lixx4266, wang6609, wnorthro, shekhar\}@umn.edu
}

\begin{document}
\frenchspacing

\maketitle
\begin{abstract}
Given an on-board diagnostics (OBD) dataset and a physics-based emissions prediction model, this paper aims to develop an accurate and computational-efficient AI (Artificial Intelligence) method that predicts vehicle emissions. The problem is of societal importance because vehicular emissions lead to climate change and impact human health. This problem is challenging because the OBD data does not contain enough parameters needed by high-order physics models. Conversely, related work has shown that low-order physics models have poor predictive accuracy when using available OBD data. This paper uses a divergent window co-occurrence pattern detection method to develop a spatiotemporal variability-aware AI model for predicting emission values from the OBD datasets. We conducted a case-study using real-world OBD data from a local public transportation agency. Results show that the proposed AI method has approximately $65\%$ improved predictive accuracy than a non-AI low-order physics model and is approximately $35\%$ more accurate than a baseline model.
\end{abstract}
\section{Introduction}

%problem definition
On-board diagnostics (OBD) data is multi-attribute trajectory data obtained from sensors in vehicles. It contains time series of engine and vehicle performance parameters. Guided by a low-order combustion-physics-based model, this paper aims to develop an OBD-data-driven AI model to predict vehicle emissions values.

%motivation
This problem is of significant societal importance because transportation is the biggest worldwide contributor to greenhouse gases such as $CO_{2}$(Carbon dioxide) and toxic gases like $NO_{x}$ (Oxides of Nitrogen like $NO$, $NO_{2}$, etc.). These emissions impair people's health \cite{article} and global climate. An understanding of vehicle and engine behavior in the real world is essential for tracking and eventually mitigating these emissions by aiding the design of cleaner and more efficient vehicle systems.

%challenges
This problem is challenging because the processes by which they are produced are complex and dependent on many parameters. Traditional laboratory experiments conducted to measure emissions values are usually based on engine-specific steady-state measurements. However, data collection inside a laboratory is expensive compared to low-cost sensor data from vehicles on the road. 

\begin{figure}[h!]
    \centering
    \includegraphics[scale=0.06]{ 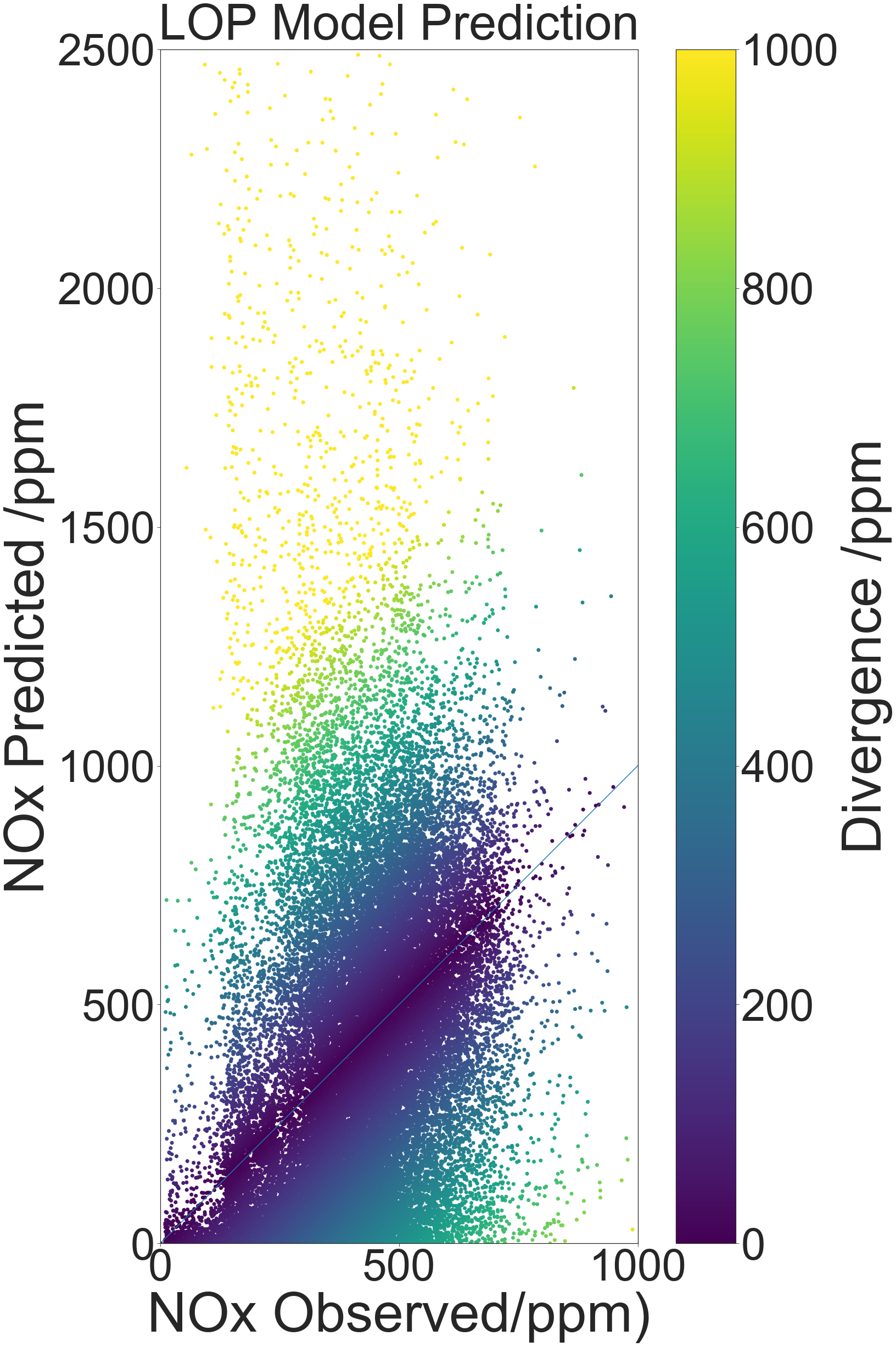}
    \caption{Comparing Observed and Predicted $NO_{x}$ values using LOP phenomenological model}
    \label{fig:noxLOP}
\end{figure}

%\subsection{Related Works}
Two types of related work are relevant: purely phenomenological methods or purely AI  approaches. \cite{HeX.2008Livc} introduced a low-order physics (LOP) model that uses a purely phenomenological method for predicting the emission index of $NO_{x}$ ($EI{-}NO_{x}$, grams of $NO_{x}$ per kilogram fuel) of a diesel engine. It assumes that $EI-NO_{x}$ depends on the intake oxygen concentration, duration of combustion, and the peak adiabatic (i.e. without heat loss) flame temperature, which was validated using engine testing observations in laboratory conditions. To show its performance in the real world, we evaluated it using an OBD dataset from transit buses in the Metro Transit (local public transportation agency), which showed that the LOP method had poor accuracy (Figure \ref{fig:noxLOP}).

% This method considers the instantaneous amount of fuel burnt during an engine cycle (intake, compression, power, and exhaust strokes) and the instantaneous heat released. The $NO_{x}$ emissions prediction portion of the model is based on an extension of the Zeldovich mechanism \cite{10.2307/44746494} for describing the rate of formation of $NO_{x}$.

% The dataset contains the measurements of 90 engine and vehicle parameters that can be used to model the working of the \textit{Cummins ISB6.7} Diesel Hybrid engine system of the transit buses in the study. Since our OBD data did not have all the parameters needed by the LOP method, we estimated the missing parameters from available data as detailed in Appendix \ref{section:physics_calculations}.  The closer the dots are to the $y=x$ line, the more accurate the prediction. The absolute values of the prediction errors are color-coded from blue to yellow, where yellow means large prediction errors.
 
An example of a purely AI approach is \cite{Obodeh2009EvaluationOA}, which evaluates the performance of an artificial neural network (ANN) on data from a laboratory test rig for an engine. However, it provides no understanding of $NO_{x}$ formation and had spurious non-physical results. Instead, engine scientists prefer an approach to predicting emissions that is interpretable using domain knowledge \cite{karpatne2018physics,7959606,li2020physics,LiYan2018Peps}.

{\bf Contributions:}
Here, we propose a novel physics-aware AI model that leverages the concepts of variability across driving scenarios, co-occurrence patterns, and a low-order combustion-physics-based model. We evaluate the proposed model using OBD data from a transit bus in the Metro Transit, Twin Cities, MN fleet. The evaluation results show that the proposed physics-aware AI model predictions are more accurate ($65\%$ lower RMSE for training data) than those of the low-order physics model for our OBD dataset.

{\bf Scope:}
The scope of this paper is limited to physics-aware, transparent, and interpretable AI models such as co-occurrence rules guided by a low-order physics-based model. Other AI models such as neural networks fall outside the scope of this work. Proprietary manufacturers' physics-based AI models are also outside the scope, due to lack of public availability of proprietary engine calibration data. This paper focuses on the prediction of $NO_{x}$ emissions from vehicles, thus other vehicular emissions are not considered.
 
% {\bf Symposium Relevance:} The paper is relevant to `Physics-guided AI to Accelerate Scientific Discovery' symposium at the AAAI Fall Symposium Series 2020 because of the synergic use of scientific principles (e.g., low-order physics) and AI methods. The work in this paper falls under two of the listed topics: (1) Discovery of physically interpretable laws from data, and (2) Hybrid constructions of physics-based and machine learning models. 

% {\bf Relation to Artificial Intelligence:} The 2019 update of the National Artificial Intelligence Research and Development Strategic Plan \cite{Kratsios2019AI} describes \textit{Data Analytics} as one of the main long-term investments that are needed to advance AI. ``\textit{Further
% investigation of multimodality machine learning is needed to enable knowledge discovery from a wide
% variety of different types of data (e.g., discrete, continuous, text, spatial, temporal, spatio-temporal,
% graphs)}." Hence, the work presented in this paper is of direct relevance to the AI community. 

% {\bf Outline:}
% The rest of the paper is organized as follows: Section 2 summarizes our baseline method for predicting emissions and Section 3 describes the proposed variability-aware AI approach to improve the performance. The experimental evaluation and discussion are given in Section 4, and Section 5 concludes the paper.

\section{Proposed Baseline Approach}
We first introduce a baseline physics-aware AI model to predict $NO_{x}$ emission values. The emission index for $NO_{x}$  $(EI-NO_{x})$ is given by a chemical kinetic equation for the extended Zeldovich Mechanism \cite{10.2307/44746494} : \begin{equation}\label{eq:ei_nox}
    EI-NO_{x} (k+ \delta)= a * T_{adiab}(k)^{b} * t_{comb}(k)^{c}
\end{equation}
% \textbf{	\ $EI{\text -}NO_{x} (k+1)= a * T_{adiab}(k)^{b} * t_{comb}(k)^{c}$}  ...Eq.1

\noindent where, $EI-NO_{x}(k+\delta)$ is the $EI-NO_{x}$ in grams $NO_{x}$ per kilogram fuel at time `k+$\delta$'; $T_{adiab}(k)$ is the adiabatic flame temperature in kelvin at time `$k$'; $t_{comb}$ is the duration of combustion in seconds at time stamp `$k$', which is approximately equal to the fuel injection duration; $a, b, c$ are constants; and $\delta$ is the time lag between the adiabatic flame temperature $T_{adiab}$ and duration of combustion $t_{comb}$ with the corresponding $NO_x$ emission index $EI-NO_x$. More details of the physics calculations are provided in the Appendix Sections of  \cite{panneerselvam2020}

\begin{figure}[h!]
    \centering
    \includegraphics[width=5cm]{ 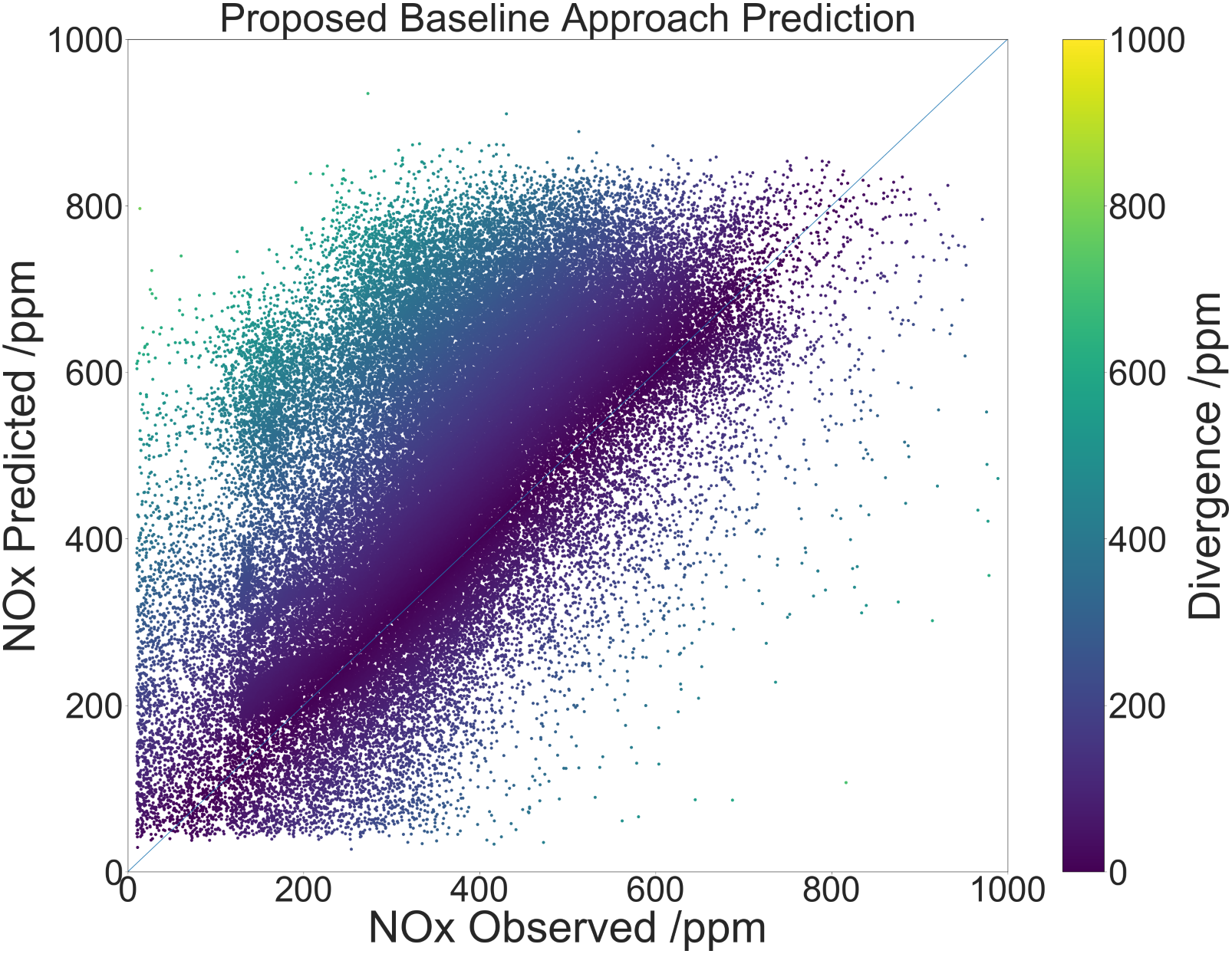}
    \caption{Comparing Observed and Predicted $NO_{x}$ values using baseline physics-aware AI model}
    \label{fig:noxbaseline}
\end{figure}

We evaluated the baseline physics-aware AI model using the same OBD dataset as the one used for the LOP method. First, we used six engine attributes (i.e. intake air flowrate(kilograms per hour), fuel consumed (kilograms per hour), rail pressure (pascal), intake pressure (pascal), intake temperature (kelvin), engine speed (revolutions per minute)) to calculate $T_{adiab}$ and $t_{comb}$ . Then, we applied a nonlinear regression method from Python Scikit-Learn package \cite{JMLR:v12:pedregosa11a} to estimate the values of $a, b$ and $c$ in Equation \ref{eq:ei_nox}. The value of $\delta = 1$ was derived using hand computation and data visualization.

Figure \ref{fig:noxbaseline} shows the $NO_{x}$ values predicted using baseline model compared with the actual values. The baseline model is an improvement over the low-order physics (LOP) model (Figure \ref{fig:noxLOP}), however, there is room for further improvement.

\section{Proposed Variability-Aware Approach}
To overcome the limitations of the baseline method, we propose a spatiotemporal (ST) variability-aware AI approach. Since one group of estimated parameters (e.g., $a, b, c$) values in Equation \ref{eq:ei_nox} does not fit all scenarios well, it may be beneficial to initially partition the data into multiple homogeneous groups, and estimate parameter values group-wise.

%cite data partitioning paper
% \cite{jiang2019spatial,dietterich2000ensemble}. 

\begin{figure}[h!]
    \centering
    \includegraphics[width=6.5cm]{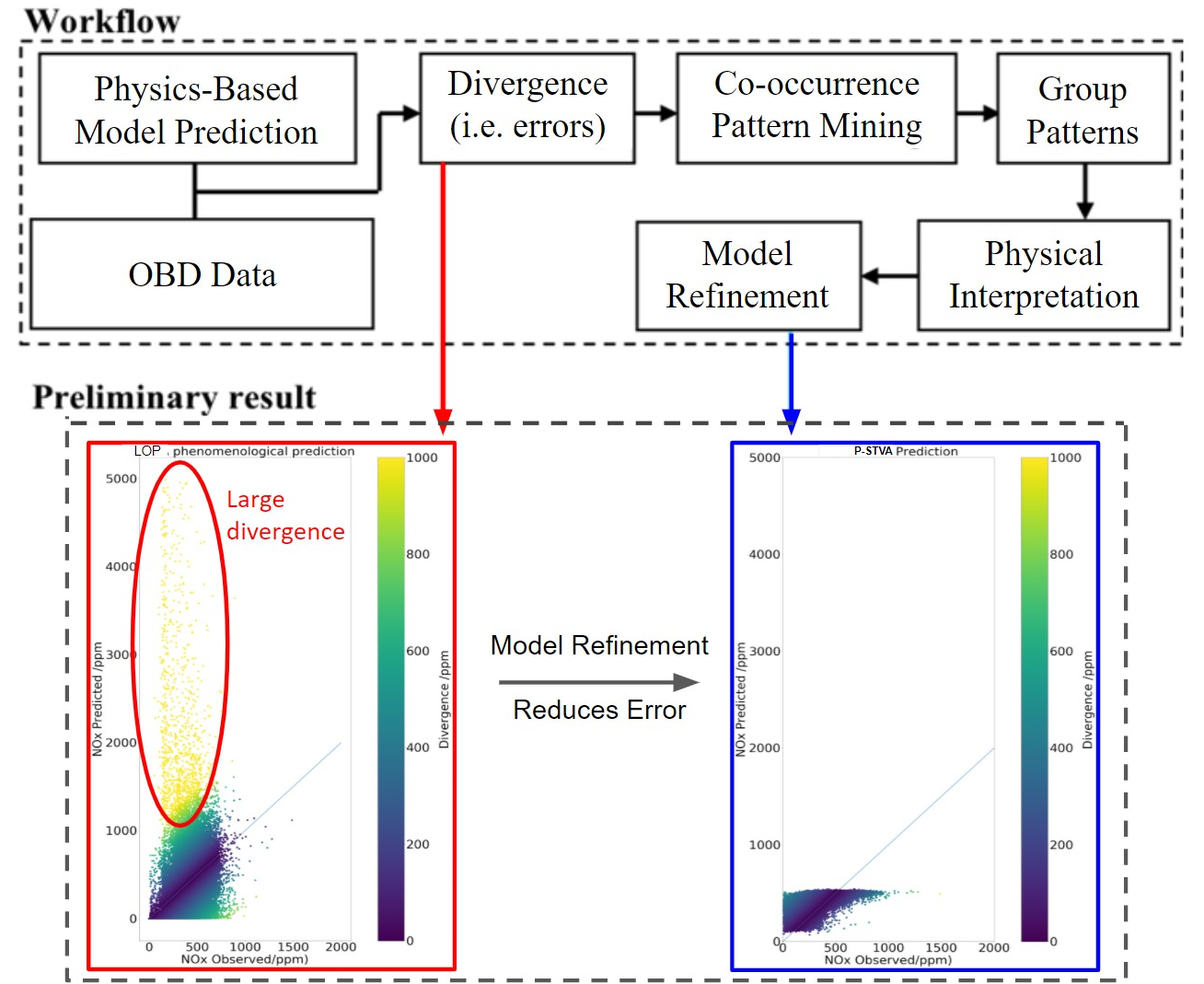}
    \caption{Proposed ST Variability-aware AI framework. The figure in the bottom half is for illustration purpose only.}
    \label{fig:framework}
\end{figure}

The top half of Figure \ref{fig:framework} shows our proposed ST variability-aware AI framework. First, we test in-coming OBD data to identify $NO_x$ emissions that diverge from the predictions made by the baseline model. We define divergence as the large (i.e. above a given threshold) absolute error between the observed and predicted $NO_{x}$ values. In general, when a vehicle exhibits divergence, there are two potential pathways for understanding the issues and improving the model: (1) using AI to improve the prediction results, or (2) using physics-based methods to develop new and refined process-based mechanistic models. 

This paper focuses on AI model refinement based on data partitioning and fitting separate models to each partition. The partitioning is based on ST correlates of divergent observations, thus we call it an ST variability-aware AI approach. This approach can potentially reduce prediction errors as illustrated in Figure \ref{fig:framework} (lower half).

A divergent window of $NO_x$ emissions refers to a period of a certain length in a time series of OBD data records within which the prediction errors of the baseline approach exceed an input threshold (\textit{summationThreshold}). A co-occurrence pattern in a time series of OBD data records is similar to a sequential association pattern \cite{Srikant1996MiningSP} except for the use of spatial statistical interest measure, i.e., temporal form of Ripley's cross-k function ($\epsilon$)  \cite{AliReem2017Dnwc,ripley1976second}. These pattern represents those subsets of engine attributes with their specific value ranges, which are present together in many divergent (time-) windows and have cross-k function values above a given threshold ($\epsilon$). Engine scientists review and group co-occurrence patterns into scenarios (for example, cold start of an engine, sudden acceleration, etc.) for physically interpreting situations where the baseline model performs poorly. 
% Table \ref{tab:cooccurrence} shows examples of co-occurrence patterns from the Metro Transit OBD dataset along with their grouping and interpretation. For example, the group "Low Vehicle Speed Condition” has two examples of co-occurrence patterns based on low wheel speed, represented as $w_0$ in a time period of three time points where $w_0$ represents the lowest value range for the wheel speed parameter. Similarly, the mined patterns with very low exhaust gas recirculation (EGR) rate and those during transient events are grouped, since the EGR rate is an important factor influencing $NO_{x}$ formation and transient events cannot be derived from stationary laboratory conditions. 

Given the co-occurrence pattern groups formed in the first step, the original OBD data is split into multiple subsets corresponding to different pattern groups. Within each subset, we use the baseline approach to calculate the values of $T_{adiab}$ and $t_{comb}$, and then estimate the parameter ($a, b, c$) values in Equation \ref{eq:ei_nox} by fitting nonlinear regression models independently. Since the scenarios when the baseline approach does not perform well are handled separately, the ST variability-aware AI model is expected to yield better predictive accuracy by lowering errors.

%explain divergent pattern
% \begin{table}
%   \begin{center}
%     \caption{Examples of divergent co-occurrence patterns}
    
%     \label{tab:cooccurrence}
%     \resizebox{8.3cm}{!}{%
    
%     \begin{tabular}{m{2.5cm}|c c}
%   	\hline
%       \textbf{Scenario} & \textbf{ Example Patterns } \\
%       \hline
%       Low Vehicle Speed Condition &   \makecell{1. Wheelspeed  $w_{0}$ $w_{0}$ $w_{0}$ \\ RailMPa $r_{2}$ $r_{2}$ $r_{2}$ \\ IntakeT $I_{9}$ $I_{9}$ $I_{9}$} &\makecell{2. Wheelspeed  $w_{0}$ $w_{0}$ $w_{0}$ \\ IntakeT $I_{9}$ $I_{9}$ $I_{9}$\\
%       Fuelconskgph $f_{2}$ $f_{2}$ $f_{2}$}\\
%       \hline
%      Low EGR Condition &   \makecell{3. Acceleration  $a_{9}$ $a_{9}$ $a_{9}$ \\ EGRkgph $g_{0}$ $g_{0}$ $g_{0}$ \\} & \makecell{4. Bkpwr  $B_{6}$ $B_{6}$ $B_{6}$ \\ EGRkgph $g_{0}$ $g_{0}$ $g_{0}$ }\\
%      \hline
%      Transient Condition &   \makecell{5. Wheelspeed  $w_{9}$ $w_{10}$ $w_{10}$ \\ Bkpwr $g_{0}$ $g_{0}$ $g_{0}$ \\} & \makecell{6. Acceleration  $a_{8}$ $a_{8}$ $a_{7}$ \\ RailMPa $r_{6}$ $r_{6}$ $r_{6}$ }\\
%       \hline
%     \end{tabular}%
    
%     }
    
%      \resizebox{3.5cm}{!}{%
%     \begin{tabular}{|c|c|}
%   	\hline
%       \textbf{Subscript} & \textbf{ Scale of values } \\
%       \hline
%       0,1 & Very low value \\
%       2,3,4 & Low value \\
%       5,6,7 & Medium Value \\
%       8,9,10 & High value \\
         
%       \hline
%     \end{tabular}%
%     }
    
%   \end{center}
  
%  \end{table}

\section{Experimental Evaluation and Discussion}
We conducted experiments to compare predictive accuracy of the proposed approaches with the low-order physics (LOP) approach detailed in Section 1 \cite{HeX.2008Livc} to address the following questions: (1) How do the predictions of the proposed approaches compare with those from the low-order physics approach? (2) How sensitive is the proposed spatiotemporal variability-aware AI approach to the number of partitions, input divergence threshold, and window length?  

% The experiment design is summarized in Figure \ref{fig:ExperimentDesign}.
% \begin{figure}[h!]
%     \centering
%     \includegraphics[width=7cm]{ExperimentDesign1.png}
%     \caption{Experimental design}
%     \label{fig:ExperimentDesign}
% \end{figure}

\textbf{Data:} The dataset used in the experiments is the Metro Transit OBD dataset that was used to evaluate the LOP approach and the proposed baseline approach in the earlier sections. It contains 99,895 data entries containing measurements of 90 engine and vehicle attributes . The OBD data was obtained from transit buses traversing 3 different routes for 16 different runs in the Minneapolis-St.Paul region. We used 8 runs for training sample and the remaining 8 runs for testing, ensuring each route is represented in both samples. More details are provided in \cite{panneerselvam2020}

\textbf{Candidate methods and metrics:} The methods evaluated in the experiments include low-order physics model (LOP), the proposed baseline (P-Base), and the proposed spatiotemporal variability-aware AI approach (P-STVA). Predictive accuracy was measured using \textbf{ $R^{2}$ } values, root mean square error (RMSE) and mean absolute error (MAE).

\begin{figure}
    \centering
    \includegraphics[width=5cm]{ 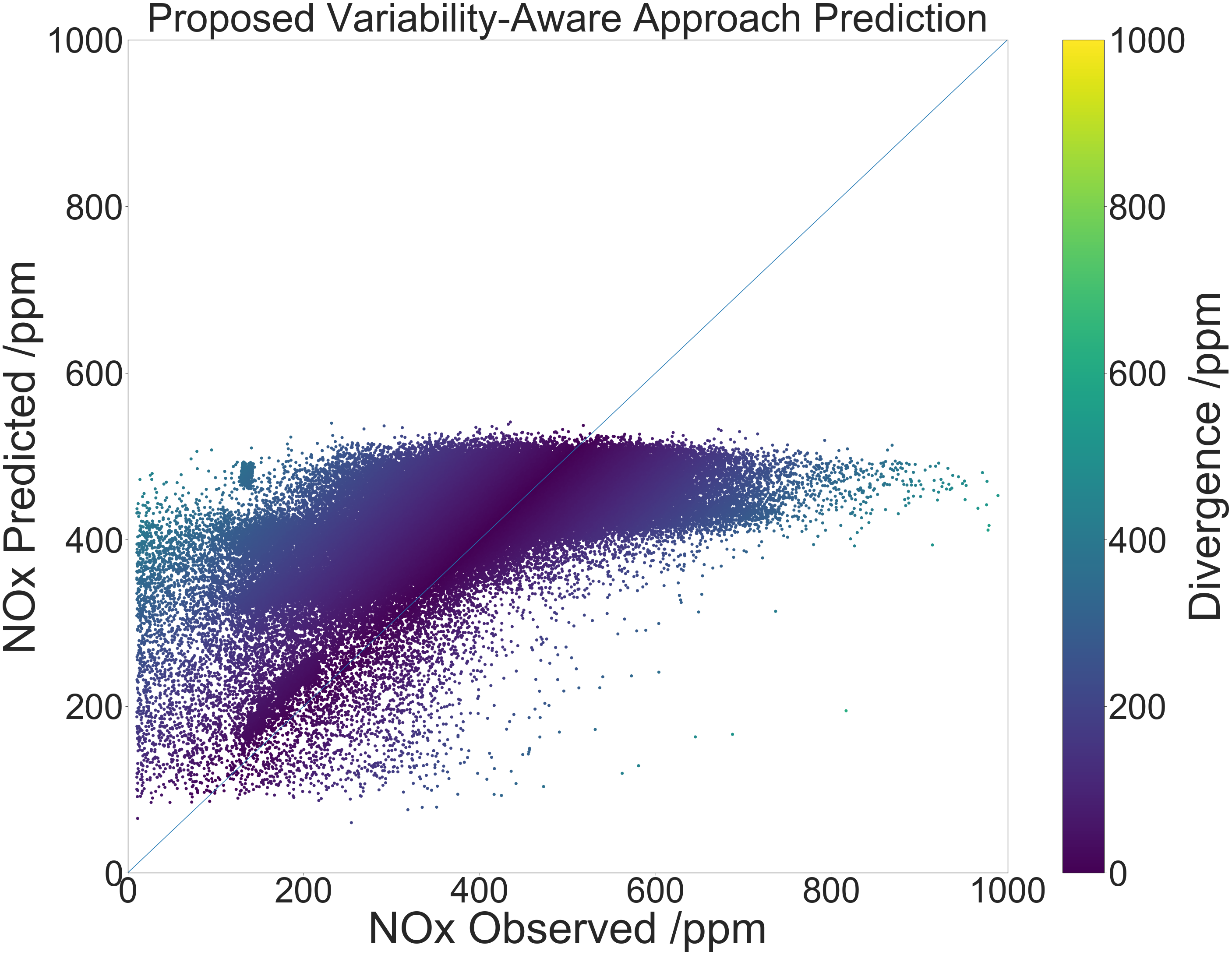}
    \caption{Refined $NO_{x}$ prediction using P-STVA method, number of partitions = 4, \textit{L} = 3 seconds, \textit{summationThreshold} = 30 ppm, \textit{minSupp} = 0.003, $\epsilon$ = 2}
    \label{fig:refined}
    \vspace{-10pt}
\end{figure}

\textbf{Experimental Results:} Figure \ref{fig:refined} shows a comparison of the refined $NO_x$ prediction using the P-STVA method with the observed $NO_x$ values in the training data. Compared with Figure \ref{fig:noxLOP} and \ref{fig:noxbaseline}, the dots in Figure \ref{fig:refined} are closer to the $y=x$ line, and the number of green and yellow dots in the upper-left part of Figure \ref{fig:noxbaseline} reduces dramatically, which indicates improved predictive accuracy.

\textit{How do the predictions of the proposed approaches compare with those from the low-order physics approach?} Table \ref{tab:trainresult} and \ref{tab:testresult} summarize predictive accuracy metrics for the candidate methods on training and testing data respectively with $n=4$ and \textit{summationThreshold} = 30 ppm (parts per million). The proposed physics-aware AI methods outperformed the low-order physics model. For the training data, the P-Base method provides about $50\%$ improvement in RMSE and $35\%$ improvement in MAE when compared to the LOP method, while RMSE and MAE of the P-STVA method with $n=4$ are both around $35\%$ smaller than the P-base method. For the testing data, the P-base method provides about $50\%$ improvement in RMSE and $40\%$ improvement in MAE when compared to the LOP method, while RMSE and MAE of the P-STVA method with $n=4$ are both around $35\%$ smaller than the P-base method. 

\vspace{-10pt}
\begin{table}[h!]
  \begin{center}
    \caption{$NO_{x}$ predictive accuracy for training data}
    \label{tab:trainresult}
    \resizebox{6cm}{!}{%
    \begin{tabular}{l|c|c|c}
    \hline
      \textbf{Prediction method} & \textbf{ R\textsuperscript{2} } & \textbf{ RMSE } & {\textbf{MAE}} \\
      \hline
      LOP & 0.1264 & 371.16 & 238.87\\
      P-Base &  0.4464 & 196.39 & 155.17\\
    %   R-ADW & 0.5821 & 140.91 & 104.84 \\
    %   R-OUP & 0.5759 & 132.84 & 97.84 \\
      P-STVA n = 4  & 0.3900 & 132.60 & 102.13\\
    %   P-STVA n=20  & 0.4309 &  120.79 & 95.87\\
      \hline
    \end{tabular}%
    }
  \end{center}
  
  \begin{center}
    \caption{$NO_{x}$ predictive accuracy for testing data}
    \label{tab:testresult}
    \resizebox{6cm}{!}{%
    \begin{tabular}{l|c|c|c}
    \hline
      \textbf{Prediction method} & \textbf{ R\textsuperscript{2} } & \textbf{ RMSE } & {\textbf{MAE}} \\
      \hline
      LOP & 0.1260 & 368.67 & 238.23\\
      P-Base & 0.4607 & 183.52 & 144.69\\
    %   R-ADW & 0.5821 & 140.91 & 104.84 \\
    %   R-OUP & 0.5759 & 132.84 & 97.84 \\
      P-STVA n = 4 & 0.4769 & 117.39 & 92.99\\
    %   P-STVA n=20  & 0.4558 &  123.54 & 99.83\\
      \hline
    \end{tabular}%
    }
  \end{center}
\end{table}
\vspace{-10pt}

\begin{figure*}
    \centering
    \includegraphics[scale=0.06]{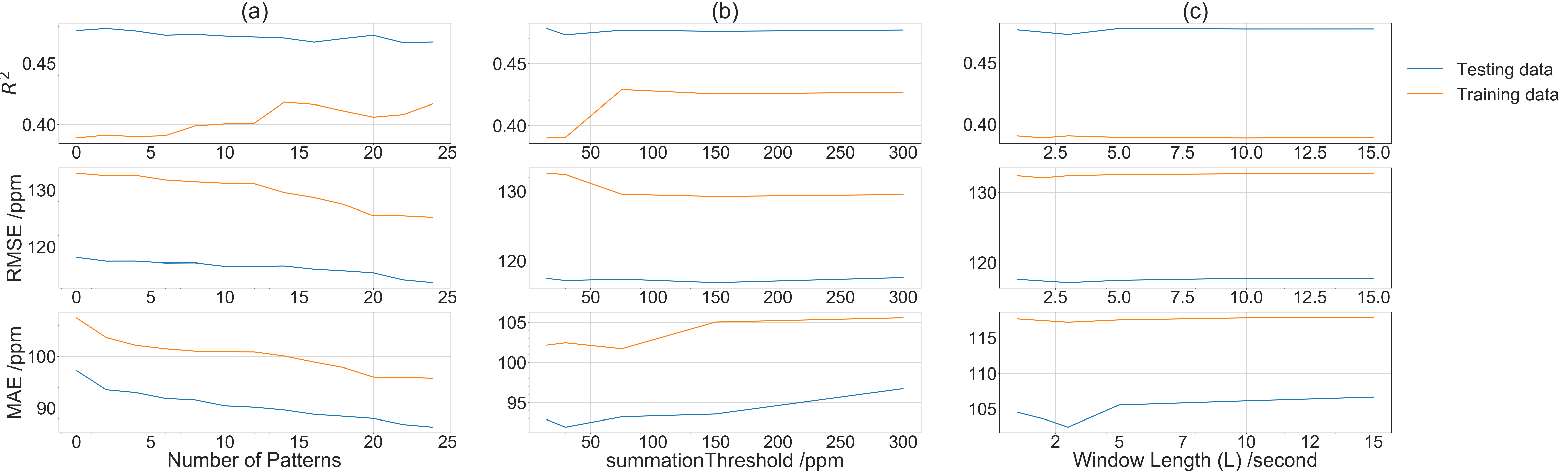}
    \caption{Sensitivity of the metrics for P-STVA to the a) number of patterns, b) \textit{summationThreshold}, and c) Window length}
    \label{fig:sensitivitypatterns}
    \vspace{-10pt}
\end{figure*}

\textit{How sensitive is the proposed spatiotemporal variability-aware AI approach to the number of partitions, input divergence threshold, and window length?
} Figures \ref{fig:sensitivitypatterns} a,b, and c show the sensitivity of predictive accuracy metrics of the P-STVA method to number of patterns $n$ (number of partitions is $n + 1$), \textit{summationThreshold}, and window Length \textit{L} respectively for training and testing data. Optimum values are found at $n=4$, \textit{summationThreshold} = 30 ppm and \textit{L} = 3 s.
\begin{table}[b]
\vspace{-10pt}
  \centering
  \caption{Four most significant divergent co-occurrence patterns in the training data}
   \resizebox{8.3cm}{!}{%
    \label{tab:patternsplotted}
    \begin{tabular}{c|c|c}
  	\hline
%   	\resizebox{6cm}{!}{%
      \textbf{Pattern} & \textbf{ Co-occurrence } & \textbf{ Scenario }\\
      \hline
       Pattern1  & EngTq: $T_{10}$ $T_{10}$ $T_{10}$ & High Engine Load \\
       \hline
       Pattern2 & EngTq: $T_{9}$ $T_{10}$ $T_{10}$ & High Load Transient \\
       \hline
       Pattern3 & \makecell{EngRPM: $R_{1}$ $R_{2}$ $R_{2}$\\ EGRkgph: $g_{4}$ $g_{4}$ $g_{4}$\\EngTq: $T_{1}$ $T_{1}$ $T_{1}$} & High Engine Idling \\ 
       \hline
       Pattern4 & \makecell{EngRPM: $R_{1}$ $R_{2}$ $R_{2}$\\ EGRkgph: $g_{4}$ $g_{4}$ $g_{4}$} & Low Engine Speed \\  \hline
    \end{tabular}%
     
    % \centering
    % \includegraphics[width=\linewidth]{ Figure_300.png}
     \resizebox{3.5cm}{!}{%
    \begin{tabular}{|c|c|}
  	\hline
      \textbf{Subscript} & \textbf{ Scale of values } \\
      \hline
      0,1 & Very low value \\
      2,3,4 & Low value \\
      5,6,7 & Medium Value \\
      8,9,10 & High value \\
      \hline
    \end{tabular}%
   }
  }
\end{table}
% Figure shows the sensitivity of the predictive accuracy to \textit{summationThreshold} for the P-STVA method for training and testing data with $n=4$ and \textit{L} = 3 seconds. The RMSE and MAE improve till \textit{summationThreshold} = 30 ppm, and then diminish.
%  \begin{figure}[h!]
%      \centering
%      \includegraphics[scale=0.1]{ Figure_28.png}
%      \caption{Sensitivity of predictive accuracy metrics of P-STVA method to \textit{summationThreshold}  }
%      \label{fig:sensitivityabs}
%  \end{figure}
% Figure  shows the sensitivity of the predictive accuracy to window length \textit{L} for P-STVA method for training and testing data with $n=4$ and \textit{summationThreshold} = 30 ppm. The RMSE and MAE improve till \textit{L} = 3 seconds, and then diminish.

%  \begin{figure*}[h!]
%      \centering
%      \includegraphics[scale=0.1]{ Figure_42.png}
%      \caption{Sensitivity of predictive accuracy metrics of P-STVA method to Window Length \textit{L } }
%      \label{fig:sensitivityL}
%  \end{figure*}
\vspace{-10pt}
\textbf{Domain interpretation of partitions:} The P-STVA method uses $n+1$ partitions including the non-divergent case (and divergent cases not covered by the patterns) and $n$ partitions of divergent cases, one for each co-occurrence pattern. Table \ref{tab:patternsplotted} shows 4 co-occurrence patterns for the case $n=4$ , \textit{summationThreshold} = 30 ppm and \textit{L} = 3 seconds along with their domain interpretation in terms of different scenarios. 

\section{Conclusions and Future Work}
We proposed a novel physics-aware AI emission prediction model and evaluated it with an on-board diagnostics dataset. The experimental evaluation shows the proposed models outperform the non-AI low-order physics model. Furthermore, the resultant models were interpreted using domain concepts as different vehicle scenarios.

In the future, we will explore other AI models such as neural networks guided by combustion physics.  We will characterize the sensitivity of the computation time of the proposed P-STVA method to parameters such as the number of partitions.  We will also investigate physics-aware AI models to predict vehicle emissions other than $NO_{x}$ as well as to predict energy use. This will assess the applicability of the Co-Occurrence Pattern Based Approach (Shown in top half of Figure \ref{fig:framework}) to other problems. Currently, engine scientists review and group co-occurrence patters into scenarios for domain interpretation. For the future work, we will explore building a machine learning model for grouping the patterns to assist human experts and reduce manual labor. 

 \section{Acknowledgments}
 This material is based upon work supported by the National Science Foundation under Grant No. 1901099. The views and opinions of authors expressed herein do not necessarily state or reflect those of the United States Government or any agency thereof. 
%  the NIH under Grant No. UL1 TR002494, KL2TR002492, and TL1 TR002493, the USDA under Grant No. 2017-51181-27222. 
 
 %and the OVPR Infrastructure Investment Initiative, Minnesota Supercomputing Institute (MSI), and Provost's Grand Challenges Exploratory Research and International Enhancements Grants at the University of Minnesota. 

\fontsize{9pt}{10pt} \selectfont
\bibliographystyle{aaai}
\bibliography{Panneer_Selvam2020-AAAI}

\end{document}